\documentclass{article}
\usepackage{template}

\usepackage[utf8]{inputenc} 
\usepackage[T1]{fontenc}    
\usepackage{url}            
\usepackage{booktabs}       
\usepackage{amsfonts}       
\usepackage{nicefrac}       
\usepackage{microtype}      
\usepackage{lipsum}
\usepackage{fancyhdr}       
\usepackage{graphicx}       
\graphicspath{{Fig/}}     
\usepackage{amsmath}
\usepackage{amssymb}
\usepackage{graphicx}
\usepackage{enumerate}
\usepackage{setspace}
\usepackage{comment}
\usepackage[utf8]{inputenc}
\usepackage[breaklinks=true]{hyperref}
\usepackage[noend]{algorithmic}
\usepackage{algorithm}
\usepackage[spanish, english]{babel}
\usepackage{url}
\usepackage{notoccite}
\usepackage{fancyhdr}
\usepackage{ragged2e} 
\usepackage[labelfont=bf, center]{caption}
\usepackage{multirow}
\usepackage{stackengine}
\usepackage{booktabs}
\usepackage[table,xcdraw]{xcolor}
\usepackage[normalem]{ulem}
\useunder{\uline}{\ul}{}
\usepackage{calrsfs}
\usepackage{textcomp}
\usepackage{bbm}
\usepackage{tikz}
\DeclareMathAlphabet{\pazocal}{OMS}{zplm}{m}{n}
\def\BibTeX{{\rm B\kern-.05em{\sc i\kern-.025em b}\kern-.08em
    T\kern-.1667em\lower.7ex\hbox{E}\kern-.125emX}}
\usetikzlibrary{shapes}
\usetikzlibrary{automata,positioning}

\usepackage{subcaption}

\usepackage{fancyhdr, hyperref}
\title{An improved tabular data generator with VAE-GMM integration}

\author{
  Patricia A. Apellániz\\
  Information Processing \\and Telecommunications Center \\
  Universidad Politécnica de Madrid \\
  Madrid\\
  \textit{patricia.alonsod@upm.es} \\
  \And
  Juan Parras\\
  Information Processing \\and Telecommunications Center \\
  Universidad Politécnica de Madrid \\
  Madrid\\
  \textit{j.parras@upm.es} \\
  \And
  Santiago Zazo\\
  Information Processing \\and Telecommunications Center \\
  Universidad Politécnica de Madrid \\
  Madrid\\
  \textit{santiago.zazo@upm.es} \\
}

\begin{document}

\maketitle

\justifying

\begin{abstract}
The rising use of machine learning in various fields requires robust methods to create synthetic tabular data that preserve key characteristics while mitigating data scarcity challenges. State-of-the-art approaches, such as CTGAN and TVAE, face difficulties with the intricate structures inherent in tabular data, which often comprise continuous and discrete features with non-Gaussian distributions. To address these limitations, we propose a novel approach based on Variational Autoencoders (VAEs) enhanced with a Bayesian Gaussian Mixture (BGM) model. Unlike other methods that alter the Gaussian prior of the VAE, our approach trains the VAE conventionally and then applies the BGM model to the learned latent space. This allows for a more accurate representation of the underlying data distribution during data generation. Moreover, our model offers enhanced flexibility by accommodating various differentiable distributions for individual features, enabling the handling of continuous and discrete data types. Thorough validation on three real-world datasets, including medically relevant ones, demonstrates significant outperformance compared to CTGAN and TVAE. Our model shows promise as a valuable tool for synthetic tabular data generation across diverse domains, particularly in healthcare.
\end{abstract}


\section{Introduction}
Despite significant advances in Machine Learning (ML) models, their efficacy relies heavily on access to extensive, high-quality training data. However, data scarcity, particularly in emerging fields such as healthcare care, impedes model development, validation, and progress in solving crucial challenges. Addressing this need for abundant and reliable data is paramount to unlocking the full potential of ML.

Recent years have witnessed a notable increase in research on synthetic data generation techniques. The past decade has seen the publication of numerous deep learning (DL)-based generative models for homogeneous data types such as images, audio, or text \cite{Radford2015UnsupervisedRL, yang2017midinet, zhang2017adversarial}. However, research \cite{arik2020tabnet} suggests that the generation of heterogeneous tabular data remains a significant challenge within DL. Tabular datasets are characterized by dense and sparse features, weaker or more complex correlations compared to spatial or semantic data. Additionally, tabular data often include missing values, outliers, and inconsistencies. Furthermore, preprocessing methods used for Deep Neural Networks (DNNs) can lead to information loss. This presents a critical challenge considering the widespread use and importance of tabular data in various fields.

Despite the aforementioned challenges, late advances in deep generative models offer promising solutions for data generation. In particular, Generative Adversarial Networks (GANs) show potential to generate synthetic tabular data. However, early vanilla GAN-based approaches \cite{medgan, tablegan} struggled with mixed data types and specific feature values, limiting their use with imbalanced datasets as demonstrated in \cite{survey}. Conditional GANs, such as \cite{cwgan} and CTGAN \cite{ctgan}, offer a more promising approach, using a conditional vector during generation to specify the desired labels. Although the state-of-the-art CTGAN handles both categorical and numerical features, it can struggle to converge with continuous data \cite{ctabgan}.

While GANs dominate the scene, autoencoders, particularly Variational Autoencoders (VAEs) \cite{kingma}, offer an alternative to generate synthetic tabular data \cite{ctgan, ma2020vaem}. In addition to CTGAN, the authors in \cite{ctgan} proposed TVAE that achieves superior performance and potentially becomes a state-of-the-art solution alongside CTGAN. However, there are limitations in TVAE's sampling process due to its Gaussian assumption, which might not hold for complex real-world data. Further exploration of alternative sampling techniques is necessary.

Building upon existing work in data generation and addressing the limitations of GANs, this paper proposes a novel VAE-based model inspired by TVAE. Our model incorporates key advancements that differentiate it from the state-of-the-art:
\begin{itemize}
    \item While TVAE assumes a Gaussian latent space for data sampling, our model employs a Gaussian Mixture model (GMM) within the VAE architecture. This avoids the potential limitation of a strictly Gaussian distribution imposed by the training process divergence term in VAEs \cite{Cremer2018InferenceSI}.
    
    \item While prior research explored non-Gaussian latent spaces in VAEs \cite{dilokthanakul2017deep}, \cite{Guo2020VariationalAW}, these approaches require modifications to the training process, increasing complexity. In contrast, our model leverages the learned latent representation after conventional VAE training and subsequently applies a GMM. Thus, we are able to generate better synthetic data while keeping the VAE training complexity, adding only the BGM after the VAE has been trained

    \item We propose a more flexible GMM compared to \cite{yang2020game} where a fixed number of components is used in a pre-trained VAE limited to non-tabular data. Our GMM learns the optimal number of components directly from the tabular data.

    \item To validate the effectiveness of our model, we performed a comprehensive evaluation using three real-world datasets, including two in the medically relevant domain, given the significance of data generation in this field. We compare our model's performance against state-of-the-art solutions like CTGAN and TVAE, employing discriminative and ML utility metrics.
\end{itemize}

\section{Methodology}
\subsection{Variational Autoencoder}
The VAE was first introduced by \cite{kingma} as a probabilistic generative model to perform Bayesian inference on datasets consisting of i.i.d. samples. The core principle lies in learning a probabilistic generative model to capture the underlying latent structure within the data. Formally, a VAE assumes a data generation process that involves two steps for each sample $\{x_i\}_{i=0}^N$: (1) sampling a latent variable $z_i$ from a prior distribution $p(z)$, which is usually assumed to be isotropic Gaussian, and (2) generating the observed data $x_i$ from a conditional distribution $p_\theta(x\vert z)$, where $\theta$ represents the model parameters. 
\begin{figure}[t]
    \begin{minipage}{\linewidth}
	\centering
        \begin{tikzpicture}[shorten >=1pt,on grid,auto] 
            \node[draw,circle,minimum width=0.5cm,minimum height=0.5cm, rotate=0, fill=gray] (z)   {z};
            \node[draw, circle,minimum width=0.5cm,minimum height=0.5cm, rotate=0] (x) [below=1.5cm of z] {x};
            \path[->] 
            (z) edge [bend left] node {$p_\theta (x|z)$} (x)
            (x) edge [bend left, dotted] node {$q_\phi (z|x)$} (z);
	\end{tikzpicture}
    \end{minipage}
     \caption{Bayesian VAE vanilla model. $z$ represents the latent variable, while $x$ denotes the observable. $p_\theta(x\vert z)$ and $q_\phi(z\vert x)$ are the generative model and the variational approximation to the unknown true posterior $p(z\vert x)$, respectively.}
    \label{fig:vae}
\end{figure}
Due to the inherent difficulty of estimating the true posterior density, VAEs employ variational methods. These methods involve defining an approximation, a variational distribution $q_\phi(z\vert x)$ parameterized by $\phi$, and optimizing its parameters to closely resemble the true posterior (Fig. \ref{fig:vae}). This optimization typically involves maximizing the Evidence Lower Bound (ELBO), which provides a lower bound on the data's marginal log-likelihood. For any sample $x_i$, the ELBO expression is:
\begin{equation}
\begin{split}
        \log p_{\theta}(x_{i})\geq-D_{KL}(q_{\phi}(z|x_{i})||p(z))\\
        +\mathbb{E}_{q_{\phi}(z|x_{i})}\left[\log p_{\theta}(x_{i}|z)\right]=\mathcal{L}(x_{i},\theta,\phi),
\end{split}
\label{eq:basic_ELBO}
\end{equation}
where $D_{KL}$ denotes the Kullback-Leibler (KL) divergence. The challenge of computing the gradient of the ELBO with respect to $\phi$ is addressed by the reparameterization trick. This technique allows ELBO estimation using samples from a simple distribution $p(\epsilon)$. In essence, a deterministic transformation $g_\phi$ is applied to the noise term $\epsilon$, resulting in the latent variable $z = g_\phi(x, \epsilon)$. This ensures that $z$ follows the variational distribution $q_\phi(z\vert x)$ while $\epsilon$ remains drawn from $p(\epsilon)$.

The VAE implementation consists of an encoder and a decoder, both implemented using DNNs. The encoder approximates the true posterior distribution $p(z|x)$ with a variational distribution $q_\phi(z|x)$, parameterized by $\phi$. The parameters of this distribution are obtained by applying a deterministic transformation $g_{\phi}$ to the input data $x$ combined with random noise $\epsilon$. The resulting latent variable $z$ is input to the decoder, which aims to reconstruct the original data by estimating the conditional distribution $p_\theta(x|z)$, parameterized by $\theta$.

\subsection{Gaussian Mixture model}
\begin{figure*}[t]
    \includegraphics[width=0.85\textwidth]{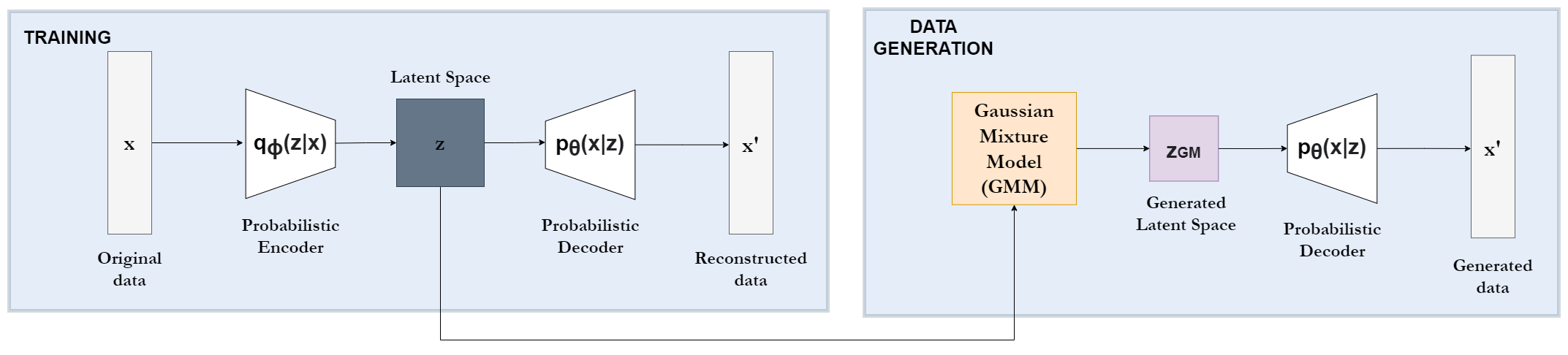}
    \centering
    \caption{Proposed model architecture. It is built on a standard VAE. After training, the latent space $z$ is modeled using a GMM. This creates a new space $z_{GM}$, which serves as the basis for generating new distribution parameters and ultimately sampling new data points.}
    \label{fig:arch}
\end{figure*}
The GMM offers a probabilistic framework for clustering data points based on Gaussian distributions. Formally, given the same dataset $X = \{x_i\}_{i=0}^N$ that contains $N$ samples with $m$ features each, the GMM assumes that the data originate from a mixture of $K$ Gaussian distributions. Each component within this mixture is characterized by its mean vector $\mu_k$, covariance matrix $\sum_k$, and mixing coefficient $\pi_k$. These parameters collectively define the shape, location, and contribution of each Gaussian component to the overall mixture. The likelihood of observing a data point $x$ as a weighted sum of the densities of individual Gaussian components is:
\begin{equation}
    p(x) = \sum_{k=1}^{K} \pi_k \pazocal{N}(x\mid \mu_k, \Sigma_k),
\end{equation}
where $\pazocal{N}(x|\mu_k, \Sigma_k)$ represents the probability density function of a multivariate Gaussian distribution with mean $\mu_k$ and covariance matrix $\sum_k$.
Estimating the model parameters, $\mu_k$, $\sum_k$, and $\pi_k$, is typically achieved by iterative optimization algorithms. The commonly employed Expectation-Maximization (EM) algorithm provides a widely used approach for this task. 

GMMs are known to be universal approximators for continuous densities \cite{kostantinos2000gaussian}, and we will rely on that property. However, GMMs face the challenge of needing $K$ as a parameter, i.e., it has to be predetermined before training. To overcome this issue, $K$ can be estimated using a Dirichlet process \cite{10.5555/1162264} using Variational Inference tools. This approach, called Bayesian Gaussian Mixture (BGM), has the advantage of automatically adjusting $K$ from the data.

\subsection{Contribution}
This work introduces a novel approach to data generation by integrating a BGM into a standard VAE architecture. This integration leverages the strengths of both models: the VAE's ability to learn a latent representation of the data $z$, and the BGM's flexibility in modeling complex distributions, potentially non-Gaussian, within this latent space. Fig. \ref{fig:arch} depicts our model. 

Previous models such as TVAE implicitly assume that the latent space $z$ aligns with the prior Gaussian distribution. Therefore, to generate new samples of $z$, sampling from the prior distribution, which is isotropic Gaussian, suffices. The reason is the presence of the KL divergence term in the loss function \ref{eq:basic_ELBO}. However, this assumption may not hold because of complex dependencies and correlations in real-world data. These factors are represented by the other term in the loss function, the marginal log-likelihood distributions. This term can push the latent space away from a Gaussian distribution. In other words, the VAE loss \ref{eq:basic_ELBO} has two components: the log-likelihood of the reconstructed data, which allows the reconstruction of samples from the latent space with high fidelity, and the KL divergence, which acts as a regularizer in the latent space. The KL divergence limits the complexity of the latent space, so it can be thought of as the allowed latent representation complexity. As it is typical to use an isotropic Gaussian as prior, due to the KL being analytical, there is an equilibrium between having a good reconstruction with a low-complexity latent space representation. In practice, the actual latent space learned need not be an isotropic Gaussian (as we will show experimentally), and thus sampling $z$ from the prior may not provide the best results in generating new samples.

Therefore, our key contribution lies in employing a BGM for the sampling process of new tabular data. Unlike TVAE, our model does not assume that the actual latent space follows the isotropic Gaussian prior: instead, we additionally model the already learned latent space $z$ as a mixture of $K$ Gaussian distributions, each characterized by its mean, covariance matrix, and mixing coefficient. Notably, if $z$ is truly an isotropic Gaussian, the BGM would still be able to effectively approximate it, as it corresponds to the case where $K=1$.

\section{Experimental analysis}
\begin{figure*}[t]
     \centering
     \begin{subfigure}[b]{0.3\textwidth}
         \centering
         \includegraphics[width=\textwidth]{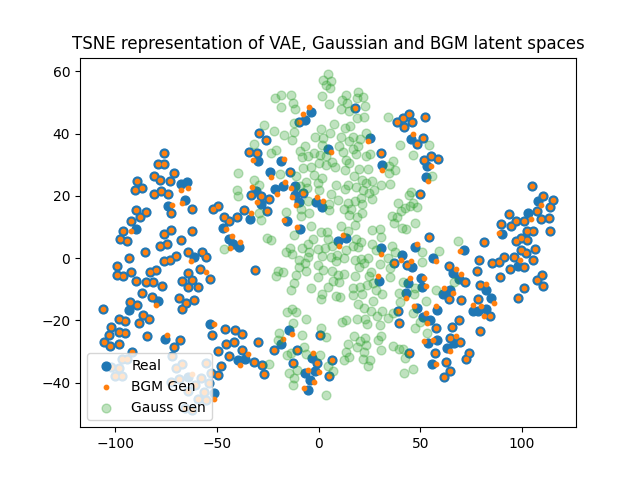}
         \caption{Adult}
     \end{subfigure}
     \hfill
     \begin{subfigure}[b]{0.3\textwidth}
         \centering
         \includegraphics[width=\textwidth]{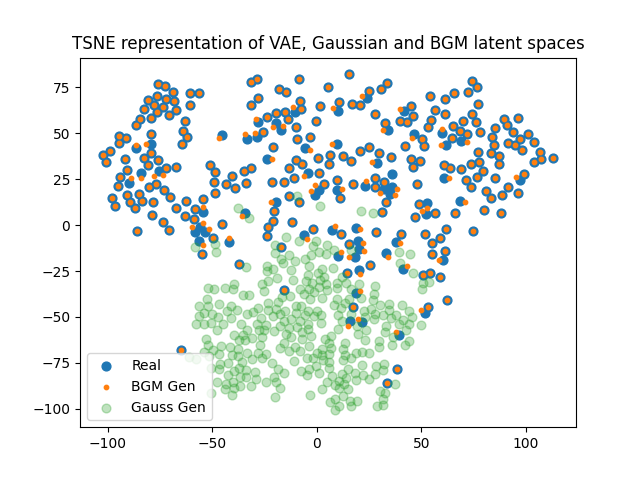}
         \caption{Metabric}
     \end{subfigure}
     \hfill
     \begin{subfigure}[b]{0.3\textwidth}
         \centering
         \includegraphics[width=\textwidth]{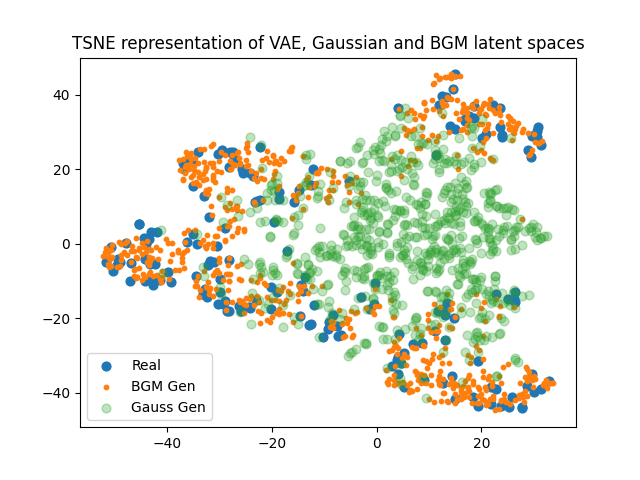}
         \caption{STD}
     \end{subfigure}
        \caption{Latent space comparison. 300 samples from each VAE, TVAE, and proposed model's latent space are shown. The BGM-modeled latent space (orange) closely aligns with $z$ (blue) compared to the TVAE's (green). This demonstrates the importance of BGM for capturing the latent space distribution and potentially leading to higher-quality generated samples. }
        \label{fig:latent_spaces}
\end{figure*}

To validate our approach, we used three diverse publicly available datasets. The first serves as a well-established benchmark commonly used in tabular data. This selection allows for comparison of our approach's performance with existing methods on a familiar task and data distribution. The remaining two datasets originate from Survival Analysis (SA) in the medical domain, where data scarcity and privacy concerns pose significant challenges. This focus aligns with the potential real-world impact of our approach, as synthetic data generation is valuable in domains where acquiring and sharing real data can be difficult or ethically problematic. Each dataset provides a unique combination of data types and target variables:
\begin{itemize}
    \item \textbf{Adult}: The Adult dataset \cite{adult} provides information from 32,561 people to predict whether a person makes more than \$50K a year. It is an extract from the 1994 Census database \cite{census}. We obtained the data from \cite{sdv} and used 10,000 for our experiments. This dataset contains 14 features with mixed data types (categorical, binary, integer).
    \item \textbf{Metabric}: The Molecular Taxonomy of Breast Cancer International Consortium (Metabric) \cite{metabric} comprises 9 clinical and genetic features from 1,904 breast cancer patients. Pre-processed data from \cite{Katzman_2018} include binary and decimal values and survival information. This selection allows us to evaluate the model's ability to handle survival data, a common target variable in medical research.
    \item \textbf{STD}: This dataset deals with Sexually Transmitted Diseases (STD) information from 877 patients. The dataset was downloaded from KMsurv \cite{std} R package. It has 23 features with categorical, integer and binary data types. The target variable focuses on survival information.
\end{itemize}

The code for our approach is publicly available on \url{https://github.com/Patricia-A-Apellaniz/vae-bgm_data_generator}. 

Our experimental setup involves two models. The VAE architecture consists of a simple encoder with a hidden layer of rectified linear units (ReLU) and a hyperbolic tangent output layer. The decoder mirrors this structure with a ReLU hidden layer, a 20\% dropout layer, and an output layer with activation functions adjusted to the covariate distributions. We fix the latent space dimensionality (number of Gaussian distributions) to 5 and the number of hidden neurons to 50. During the training of 1,000 epochs and a batch size of 500, an early stop is configured based on the validation loss. The GMM employs a Dirichlet process prior with a fixed maximum number of components set to the dimensionality of the latent space (in this case, 5) and individual general covariance matrices for each component. For training and evaluation, data are divided into 80\% training and 20\% validation sets. Due to the sensitivity of VAEs to initial conditions, 15 training runs with different seeds are performed. The final results are averaged from the 3 seeds with the best performance.

Fig. \ref{fig:latent_spaces} verifies our hypothesis that the actual latent space learned by the VAE does not need to follow the prior distribution, so our choice of BGM is justified. The figure compares the latent spaces obtained using three different approaches: (1) the original latent space from VAE, obtained by passing the dataset samples through the encoder, (2) latent space sampled using a BGM (ours), and (3) latent space sampled using the prior isotropic Gaussian (TVAE). We can clearly see that the BGM-modeled latent space aligns closely with the original VAE latent space in all three cases. In contrast, the TVAE latent space exhibits less dispersion, suggesting that the Gaussian assumption does not accurately capture the underlying distribution of the real-world data. This visually confirms our hypothesis that the latent space in real-world datasets often deviates from a Gaussian distribution, which justifies the use of BGM for a more accurate representation in the sampling process.

Let us now confirm that intuition with benchmarks on the data. Due to the lack of standardized metrics, the evaluation of data generation models requires a multifaceted approach. We use a combination of resemblance and utility-based evaluations. Resemblance is assessed through a Random Forest (RF) discriminator that measures the similarity of the joint distribution between real and synthetic data. In an ideal scenario of high model performance, the RF accuracy should be close to a random guessing rate (around $0.5$), indicating that it can not distinguish between real and synthetic data. Furthermore, we use the technique proposed in \cite{sdv} to validate the marginal distributions and the correlation between pairs of columns. Utility evaluation, tailored to the type of task, involves the Cox Proportional Hazards (Coxph) model \cite{cph} and the Concordance Index (C-index) \cite{antolini} for SA, and the RF discriminator for classification tasks. This evaluation considers two scenarios: training and testing on real data (benchmark) and training on synthetic data with real data testing. This assessment ensures that the generated data exhibit high resemblance (capturing feature distributions and correlations) and utility (supporting robust statistical inference). To benchmark our model's performance, we compared it with CTGAN and TVAE, the current state-of-the-art models for synthetic data generation. For a coherent comparison, TVAE and our model are implemented in the same way, but we add the BGM sampling to ours. All models were evaluated using the same methodology for a fair and consistent comparison.

\begin{table}[h!]
\caption{Resemblance evaluation using RF}
    \centering
        \begin{tabular}{cccc}
            \hline
            \textbf{Dataset} & \textbf{CTGAN}   & \textbf{TVAE} & \textbf{Our model}\\
            \hline
                Adult   & 0.75 (0.74, 0.76)& 0.78 (0.74, 0.81) & \textbf{0.68 (0.64, 71)} \\
                  Metabric &  \textbf{0.73 (0.70, 0.76)}&  0.77 (0.74, 0.80)& \textbf{0.67 (0.62, 0.71)}\\
                  STD &    0.94 (0.91, 0.96)& 0.77 (0.70, 0.82)&  \textbf{0.64 (0.57, 0.70)}\\
            \hline
             \multicolumn{4}{c}{Accuracy and 99\% CI. Lower is better. Best values are in bold. }
        \end{tabular}
    \label{tab:rf_res_results}
\end{table}

\begin{table}[h!]
\caption{Resemblance evaluation using column analysis \cite{sdv}}
    \centering
        \begin{tabular}{cccc}
            \hline
            \textbf{Dataset} & \textbf{CTGAN}   & \textbf{TVAE} & \textbf{Our model}\\
            \hline
                Adult   & 0.87 &   0.87 & \textbf{0.93}\\
                 Metabric  & 0.89 & 0.88 & \textbf{0.92} \\
                STD & 0.86  & 0.87 &  \textbf{0.95}\\
            \hline
            \multicolumn{4}{c}{Higher is better: a score of $1$ means that the patterns captured for real}\\
            \multicolumn{4}{c}{and synthetic data are exactly the same. Best values are in bold.}
        \end{tabular}
    \label{tab:sdv_res_results}
\end{table}

The resemblance results of the proposed model compared to CTGAN and the VAE-based Gaussian sampling model can be seen in Tables \ref{tab:rf_res_results} and \ref{tab:sdv_res_results}. Table \ref{tab:rf_res_results} reports the accuracy and its Confidence Intervals (CI) obtained by the RF for each model. For VAE-based models, the table shows the average of the three best seeds and the CIs are calculated based on the ones obtained for each seed. The results show that our model consistently outperformed the other two models by a significant margin in all datasets, indicating superior generative capabilities and robustness. Furthermore, the results of the column analysis in Table \ref{tab:sdv_res_results} confirm the superiority of our model in terms of similarity to ground truth values.  

\begin{table*}[h!]
\caption{ML task-related results comparing CTGAN, TVAE, and our model}
    \centering
        \begin{tabular}{ccccc}
            \hline
            \textbf{Dataset} & \textbf{Benchmark} &  \textbf{CTGAN}   & \textbf{TVAE} & \textbf{Our approach}\\
            \hline
                Adult   & 0.80 (0.79, 0.82) & 0.79 (0.77, 0.80) &  0.80 (0.77, 0.82) & 0.79 (0.76, 0.81) \\
                 Metabric  & 0.58 (0.53, 0.63) & 0.57 (0.52, 0.62) & 0.60 (0.54, 0.65)& 0.60 (0.54, 0.65) \\
                STD & 0.64 (0.57, 0.70) & 0.54 (0.47, 0.61) & 0.54 (0.46, 0.60)& 0.65 (0.55, 0.75) \\
            \hline
             \multicolumn{5}{c}{Accuracy results for Adult. C-index results for Metabric and STD. Note that all the methods tested}\\
             \multicolumn{5}{c}{provide useful data for the tasks tested, as the results fall in the CI of the benchmark,}
        \end{tabular}
    \label{tab:utilitiy_results}
\end{table*}

Table \ref{tab:utilitiy_results} presents the validation results for each ML task based on the dataset. For the Adult dataset, the accuracy is reported, as it is a classification dataset, while the C-index metric is used for the other two datasets. Similarly to the previous evaluation, we present the average value for the three best performing random seeds in the case of TVAE and our proposed method. The ``Real-Real" value represents the benchmark. The benchmark results are the upper bound in performance, where training and testing are performed using real data. Then, for each generative model, we also show the results for these ML tasks training with synthetic data and testing with real data to assess whether the generated data are useful for the intended purpose of the dataset. The results in this validation stage indicate that, in general, the performance metrics obtained using data generated by each model (TVAE and ours) are comparable to the results obtained using real data. This suggests that the generated data exhibit sufficient utility and quality for practical ML applications.

\section{Conclusions}
\label{sec:concl}
This work proposes a novel approach to generate synthetic tabular data by integrating a BGM into a VAE architecture.  Our experiments demonstrate that the proposed model outperforms the state-of-the-art models CTGAN and TVAE on several validation criteria. This superior performance can be attributed to two key strengths. First, our model effectively captures the diverse distributions present in the data, both at the individual feature level (marginal) and across features (joint). Unlike many existing models that struggle with mixed data types, our approach handles various data types effectively. Second, the incorporation of the BGM enables high-quality sampling from an approximation of the VAE's latent space. This is crucial, as TVAE's assumption of a Gaussian latent space can lead to worse performance with real-world data that often deviates from this distribution. Looking ahead, several promising lines for future research exist. One direction involves exploring the privacy implications of data generation, particularly in sensitive domains like healthcare. Analyzing privacy concerns and developing secure methods to share synthetic data in medical settings is a valuable area of investigation. Furthermore, we propose investigating novel federated learning strategies based on synthetic data sharing rather than trained model information. This approach could improve collaboration and knowledge sharing among entities while protecting sensitive information.

\section*{Acknowledgment}
This research was supported by GenoMed4All and SYNTHEMA projects. Both have received funding from the European Union’s Horizon 2020 research and innovation program under grant agreement No 101017549 and 101095530, respectively. The authors declare that they have no known competing financial interests or personal relationships that could have appeared to influence the work reported in this paper.

\bibliographystyle{unsrt}  
\bibliography{main}  

\end{document}